\documentclass[final,1p,times]{elsarticle}

\graphicspath{{fig/}} 

\journal{Pattern Recognition}

\usepackage{lineno}
\usepackage{graphicx}
\usepackage{float}
\usepackage{epstopdf}
\usepackage{amsmath,amssymb,amsfonts}
\usepackage{algorithm}
\usepackage{algpseudocode}
\usepackage{textcomp}
\usepackage{xcolor}
\usepackage{array}
\usepackage{balance}
\usepackage{indentfirst}
\usepackage{multirow}
\usepackage{booktabs}
\biboptions{sort&compress}
\usepackage{lscape}
\usepackage{tikz}
\usepackage{wrapfig}
\usepackage{mathrsfs}
\usepackage{subcaption}

\begin{document}

\begin{frontmatter}

\title{On the Adversarial Robustness of Generative Autoencoders in the Latent Space}
\author[1]{Mingfei Lu}
\author[1]{Badong Chen}

\affiliation[1]{organization={National Key Laboratory of Human-Machine Hybrid 
Augmented Intelligence, National Engineering Research Center for Visual Information and Applications, and Institute of Artificial Intelligence and Robotics, Xi'an Jiaotong University},
            addressline={}, 
            city={Xi'an},
            postcode={710049}, 
            state={},
            country={China}}

\begin{abstract}
    The generative autoencoders, such as the variational autoencoders or the adversarial autoencoders, have achieved great success in lots of real-world applications, including image generation, and signal communication. 
    However, little concern has been devoted to their robustness during practical deployment. 
    Due to the probabilistic latent structure, variational autoencoders (VAEs) may confront problems such as a mismatch between the posterior distribution of the latent and real data manifold, or discontinuity in the posterior distribution of the latent. 
    This leaves a back door for malicious attackers to collapse VAEs from the latent space, especially in scenarios where the encoder and decoder are used separately, such as communication and compressed sensing. 
    In this work, we provide the first study on the adversarial robustness of generative autoencoders in the latent space. 
    Specifically, we empirically demonstrate the latent vulnerability of popular generative autoencoders through attacks in the latent space. 
    We also evaluate the difference between variational autoencoders and their deterministic variants and observe that the latter performs better in latent robustness. 
    Meanwhile, we identify a potential trade-off between the adversarial robustness and the degree of the disentanglement of the latent codes.  
    Additionally, we also verify the feasibility of improvement for the latent robustness of VAEs through adversarial training. 
    In summary, we suggest concerning the adversarial latent robustness of the generative autoencoders, analyze several robustness-relative issues, and give some insights into a series of key challenges.
\end{abstract}


\end{frontmatter}


\section{Introduction}\label{intro}
    As one of the most successful deep unsupervised representation learning models, variational autoencoders (VAEs)~\cite{kingma2013auto,rezende2014stochastic} and their deterministic variants (such as the adversarial autoencoders~\cite{makhzani2015adversarial} and the regularized autoencoders~\cite{ghosh2020variational}) have been used in many domains such as computer vision~\cite{
    zhou2018variational, liu2021variational}, natural language processing~\cite{
    jang2019recurrent, semeniuta2017hybrid}, time series~\cite{
    li2020anomaly, jin2022pfvae}. 
	By taking advantage of the prior distribution hypothesis and the re-parameterize trick for the latent representation, VAEs outperform the classic autoencoders that are trained by minimizing reconstruction error from two perspectives: 
	(a) it helps to make smooth interpolations, which means VAEs can be used as generative models to sample from the latent space and make new reasonable examples with high quality~\cite{kingma2013auto}. 
	(b) it provides more robustness against input perturbations, particularly those originating from adversarial attacks~\cite{schott2019towards, willetts2020improving}.
	
\begin{figure}[tbp]
		\centering
		\subfloat[]{\centering\includegraphics[width=0.45\linewidth]{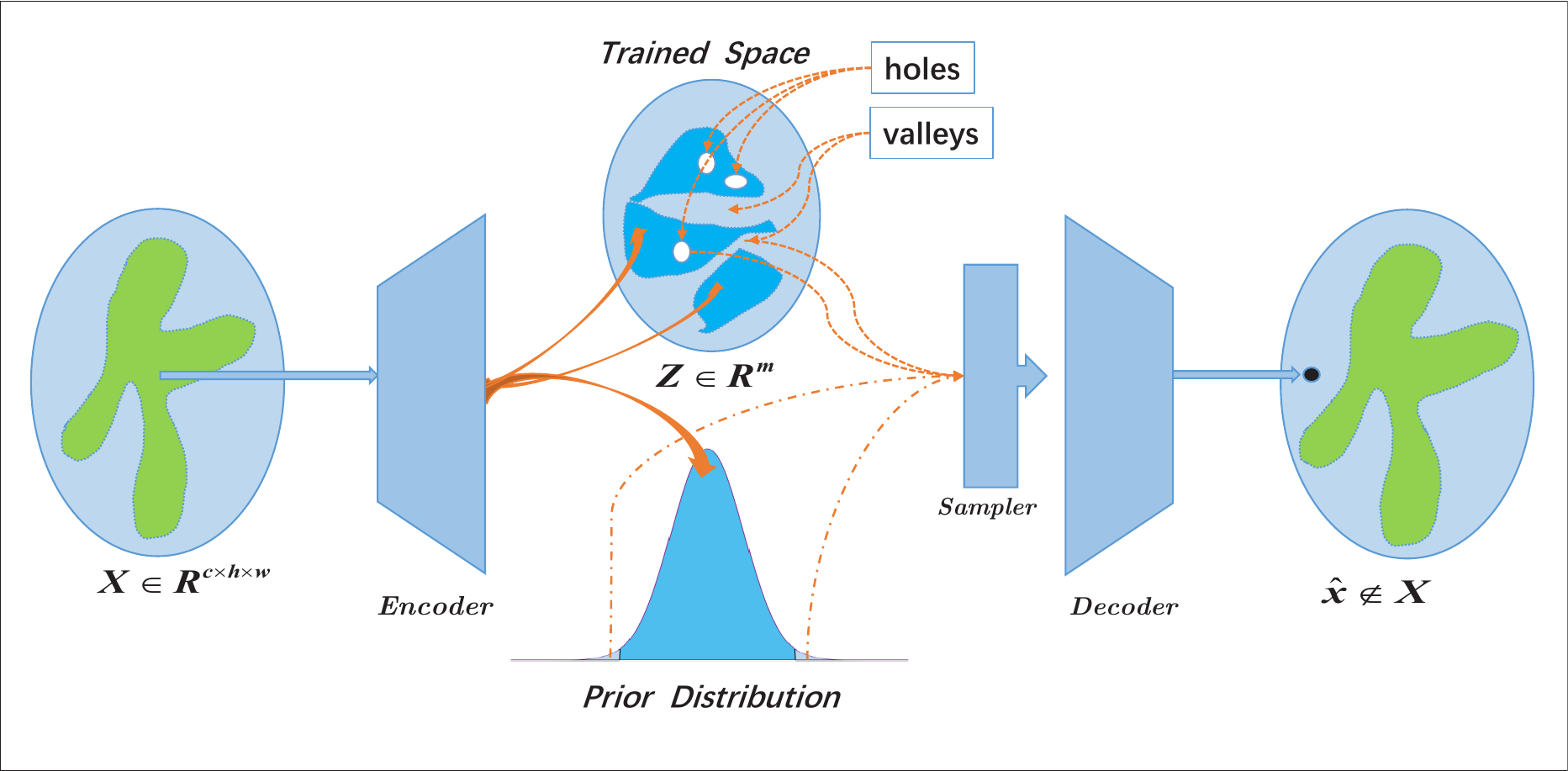 } \label{Fig:valey_holes_vae}}
		\hfill
		\subfloat[]{
                        \raisebox{0.18\height}{\centering\includegraphics[width=0.45\linewidth]{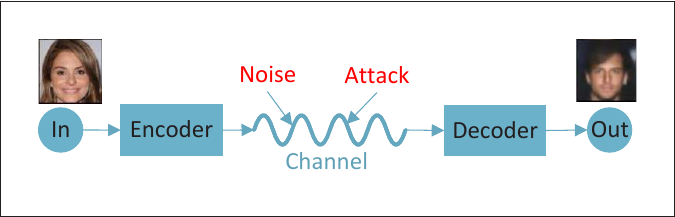 } }\label{Fig:Attack_scenarios}}
		\caption{Illustration on the vulnerability of the generative autoencoders. (a) gives a schematic illustration of reasons for VAE to get bad reconstructions or generations in principle. (b) presents the case where autoencoders are used in communication systems. When the compressed representation (latent code) of an image is transmitted from the encoder (transmitter) to the decoder (receiver), noise or intervention can be poured into the channel and change the latent, leading the decoder to reconstruct a fake image.}
		\label{Fig:Attack_illustration}
	\end{figure}

	Nevertheless, several aspects of the traditional VAE framework prevent it from trustworthy reconstruction or generating new data. 
	On the one hand, insufficiency of the training data may cause holes or valleys in the latent space~\cite{leeb2021interventional,rezende2018taming}, as illustrated in Figure~\ref{Fig:valey_holes_vae}, from where sampling a latent may lead to bad or even invalid reconstruction or generation.
    Meanwhile, a latent from the low-density area of the prior distribution also tends to produce a sample with low quality in high probability.
	On the other hand, VAEs enforce a global structure in the latent space by fitting a prior distribution that may not match the true data manifold. 
    This model mismatch can result in less accurate generative modeling of the data~\cite{connor2021variational,lucas2019understanding}. 
	
	
	Note that, the above-mentioned limitations are mostly related to the latent space of generative autoencoders. 
	In this sense, from a security perspective, the vulnerability of the generative autoencoders in the latent space may provide an easy opportunity for attackers who aim to deteriorate the reconstruction of those autoencoders (especially in a communication scenario).  
	Moreover, in practical scenarios like communication or compressed sensing~\cite{
 ,doi2011characterization,liu2022opening} as depicted in Figure~\ref{Fig:Attack_scenarios}, the encoder and decoder of an autoencoder are used separately hence the latent transmitting channel is at risk of physical interference or attack~\cite{kos2018adversarial}.
	Motivated by the above facts, we systematically, for the first time, investigate the adversarial robustness of generative autoencoders in the latent space.
	
    The adversarial robustness of generative autoencoders has been extensively investigated~\cite{
    ghosh2019resisting, sun2020type, kuzina2021diagnosing}. 
    However, most of existing studies focus on robustness against adversarial inputs, while little research has been done on the latent counterpart. 
	G. Osada et al. propose a latent space virtual adversarial training algorithm, which injects perturbation in the latent space and aims to generate input samples with more adverse-effective regularization~\cite{osada2020regularization}.
	In~\cite{yu2021lafeat}, Yu et al. point out that latent features in such input-perturbation-robust models are surprisingly susceptible to adversarial attacks. 
	Through harnessing latent features, they formulate a unified $\ell_{\infty}-$norm white-box attack algorithm with a stronger adversarial effect.
	Park et al. introduce a single-step latent adversarial training method~\cite{park2021reliably}, which leverages the gradients of latent representation as to the latent adversarial perturbation.

    It is worth note that our motivation and study in this work is totally different from ~\cite{osada2020regularization,yu2021lafeat,park2021reliably}. 
    First, we study the adversarial robustness of generative models under an autoencoder framework, rather than a discriminative models used mostly for classification. 
    Second, our study is not only targeted for developing an advanced adversarial training method to improve robustness. 
    Rather, we aim to warn practitioners of the vulnerability of autoencoders in the latent space and provide several insights with respect to both variational and deterministic autoencoders (DAE). 
    As a by product, we also demonstrated that the latent robustness of VAE models can be improved by adversarial training.
	
	
	We start the research with attack experiments to show the latent vulnerability on well-trained VAE models based on the MNIST
 , FasionMNIST
 , and CelebA
  datasets.
	Next, experiments are conducted to investigate the difference in adversarial latent robustness between VAEs and DAEs. 
	This involves a key question: whether VAE or DAE is more robust to attacks and potential for safe practical applications. 
	Another concern of ours is the relation between adversarial robustness (in latent space) and the degree of disentanglement (of latent representations). 
	It is well-known that there exists a trade-off between the reconstruction accuracy and disentangling strength for disentangling VAE such as $\beta$-VAE~\cite{higgins2016beta} and $\beta$-TCVAE~\cite{chen2018isolating}, which motivates us to consider the (possible) existence of other trade-off factors. 
	Comparison attack experiments with different $\beta$ are conducted to further reveal the mystery of whether there are trade-offs among the reconstruction accuracy, disentangling strength, and latent robustness.
	
	Our contributions are summarized below:
	\begin{itemize}
		\item Proposal of the adversarial robustness problem for generative autoencoders in the latent space (in Section~\ref{sec_problem_definition});
		\item Demonstration of the vulnerability to adversarial latent and the potential to promote latent robustness through adversarial training (in Section~\ref{sec_Attack_experiments});
		\item Investigation of the difference in latent adversarial robustness between VAEs and DAEs, and an insightful finding that deterministic autoencoders show more robustness in the latent space (in Section~\ref{method: part-II});
		\item Analysis of the trade-off between latent robustness and the disentanglement of the latent representations (in Section~\ref{method: part-III}).
	\end{itemize}
	
\section{Preliminaries and related work}\label{Preliminaries}
	
	\subsection{Variational Autoencoders}\label{VAE_intro}
\begin{figure*}[tbp]
		\centering
		\includegraphics[width=0.85\linewidth]{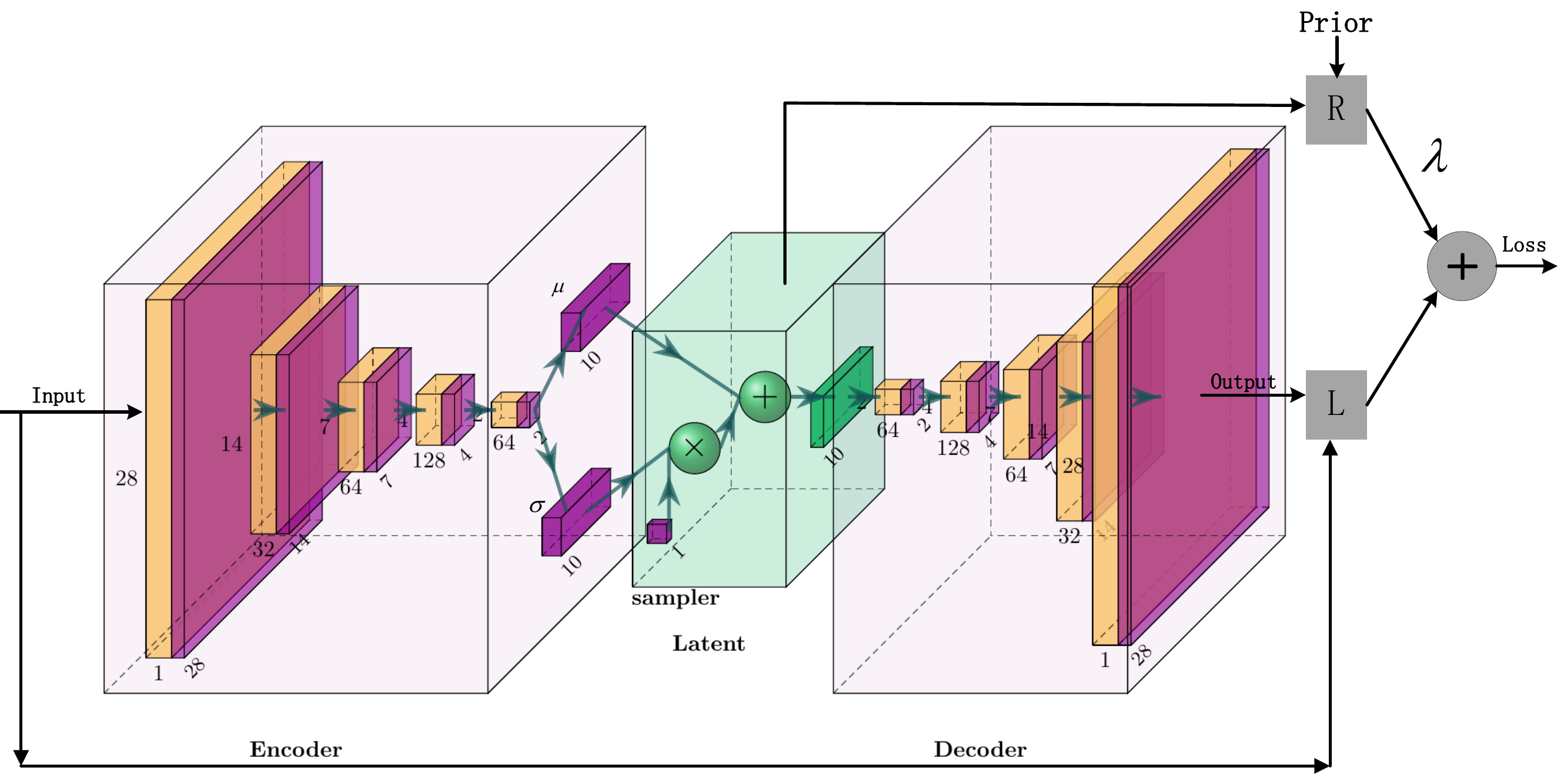}
		\caption{Structure of the VAE models. 
    		The overall loss function is consisted of the reconstruction error part $\mathbf{L}$ which is MSE if choosing normal distribution as the prior, and the regularization part $\mathbf{R}$ like KL-divergence, MMD, SWT and so on. 
		}
		\label{Fig:vae_framework}
	\end{figure*}

	In their seminal work~\cite{kingma2013auto,rezende2014stochastic}, Kingma $\&$ Rezende et al. introduced the variational autoencoder which has attracted much research interest, and become one of the most popular generative models used so far~\cite{tschannen2018recent}.
	The general framework of a VAE model is shown as in Figure~\ref{Fig:vae_framework}. 
	The \textbf{encoder} $f_{enc}(x)$ is a network mapping a high-dimensional input representation $x$ into a lower-dimensional (compressed) \textbf{latent representation} $z$. 
	And the \textbf{decoder} $f_{dec}(z)$ is a mirror network of the encoder, mapping the latent representation back to a high-dimensional output $\hat{x}$.
		
	The VAE model provides a very revolutionary idea of having neural networks learn the distribution rather than the features of the data only. 
	By applying a prior distribution hypothesis with an explicit density function for latent $Z$ and pursuing the maximum log-likelihood for the posterior distribution of the data, they derive the variational/evidence lower bound (ELBO) and then train the model. 
    Nevertheless, the ELBO objective ensures to minimize the reconstruction error and the data distribution hypothesis fitting error simultaneously:
	\begin{equation}\label{eq:Objective}
	\mathcal{L}_{\text{ELBO}} =  {D_{\text{KL}}}\left[ {q\left( {\left. z \right|x} \right)\left\| {p\left( z \right)} \right.} \right] - {E_{q\left( {\left. z \right|x} \right)}}\left[ {\log p\left( {\left. x \right|z} \right)} \right].
	\end{equation}
	The first term is the Kullback–Leibler divergence between the learned approximation $q\left( {\left. z \right|x} \right)$ to the true posterior distribution and the prior distribution of the latent representation $z$, and the second term ${E_{q\left( {\left. z \right|x} \right)}}\left[ {\log p\left( {\left. x \right|z} \right)} \right]$ denotes the loss of reconstruction $\hat{x}$ from the original input $x$. 
	In view of the objective function, $D_{\text{KL}}$ is a regularization term which quantifies the mismatch between the learned posterior distribution and the prior. 
	In practice, the KL divergence can also be replaced with the maximum mean discrepancy (MMD)~\cite{gretton2012kernel} and the cauchy-schwarz (CS) divergence~\citep{principe2010information} for more flexible latent prior, beyond just an isotropic Gaussian.
	
	Notice that the encoder does not produce a latent representation directly but the corresponding parameters for the probability density function (PDF) of the prior distribution. 
	This is the specific and most important difference between VAEs and the traditional autoencoders. 
	Then, there comes the \ \emph{re-parameterize trick} $z = \mu  + \sigma  \odot \zeta$, where $\zeta$ is randomly sampled from the prior normal distribution. 
	With such a trick, the overall framework is able to be optimized using the backward-propagation process.

	In this work, we conduct studies with the above original VAE model framework and two simple variants. 
	Referring to the understanding of the VAE loss-function in~\cite{ITL_VAE2016,yu2019understanding} where the framework is illustrated in Figure~\ref{Fig:vae_framework}, we replace the regularization term of Eq~(\ref{eq:Objective}) with MMD and SWT~\cite{jin2020adversarial}, thus achieve the MMD-VAE and SWT-VAE.  
	MMD is the non-parametric kernel two-sample test metric proposed by Gretton et al.. 
	SWT (Shapiro-Wilk Test) is a parametric distributional testing method for Normal distribution~\cite{royston1992approximating}, and we use its extension proposed in \cite{jin2020adversarial}.
    The corresponding loss functions are:
	\begin{equation}\label{eq:objective-mmd}
		{L_{ELBO-MMD}} = MMD\left[ {\left. z \right|x,{z_s}} \right] - {E_{q\left( {\left. z \right|x} \right)}}\left[ {\log p\left( {\left. x \right|z} \right)} \right],
	\end{equation}
	and
	\begin{equation}\label{eq:objective-swt}
		{L_{ELBO-SWT}} = 1 - W\left( {\left. z \right|x} \right) - {E_{q\left( {\left. z \right|x} \right)}}\left[ {\log p\left( {\left. x \right|z} \right)} \right],
	\end{equation}
	where ${z_s} \sim \mathcal{N}\left( {0,{I^{m \times d}}} \right)$ are random sampled from normal distribution, $m$ denotes the batch size, $d$ is the dimension of the latent code, and $W$ can be calculated with the method proposed in~\cite{jin2020adversarial}. 

	The original VAE model is also known as Vanilla-VAE, the objective of which employs the KL-divergence for regularization as in Eq.~(\ref{eq:Objective}). Therefore, we use the terms ``Vanilla-VAE'' and  ``KLD-VAE'' interchangeably.
\subsection{Deterministic Generative Autoencoders}\label{section-DAE}
    Traditional deep autoencoders tend to learn a trivial identity function and thus copy the input to the output, instead of picking up the underlying patterns and characteristics of the data distribution to generate new examples~\cite{oussidi2018deep}.
    VAEs bring auto-encoding into the generative era with theoretical attractiveness besides a pretty framework.
    However, they suffer from the posterior collapse problem hence motivating many studies turning back to deterministic autoencoders.
    
    
    RAE~\cite{ghosh2020variational} fixes the variance of the inferred Gaussian approximate posterior distribution as a hyper-parameter, and substitutes the stochastic encoder by injecting noise into the input of a deterministic decoder. 
    Ding et al. improve the RAE and propose the SCVG to learn the variance of the approximate Gaussian posterior distribution in a semi-deterministic manner by aggregating inferred mean vectors from other connected nodes via graph convolution operation~\cite{ding2021semi}. 
    In~\cite{wu2019couple}, the authors couple the VAE model with a deterministic network sharing the same structure but optimized with the reconstruction loss without regularization for latent distribution. 
    The DD-VAE proposed in~\cite{polykovskiy2020deterministic} employs a variational encoder but deterministic decoder.
    A family of generative models named Exemplar VAE bridges the gap between parametric and non-parametric, exemplar based generative models~\cite{norouzi2020exemplar}. 
    
    All the above works put efforts into changing the prior hypothesis and the corresponding sampling procedure for data or the framework of the original VAE, through which they achieve deterministic autoencoders.
    There is another way of obtaining deterministic generative autoencoders just by rethinking the way of generating latent variables or the organization of ELBO of the VAEs.
    
    Remember in mind that the ELBO objective defined in~Eq.(\ref{eq:Objective}) consists of two parts. 
    The second term exhibits a mean squared error (MSE) with L2 regularization on $\mu_q(x)$, which helps to reduce reconstruction loss. 
    While the first term, representing KL-divergence between the posterior data distribution and its prior one, works to fitting the data distribution.
    Obviously, the implementation of the KL-divergence is the real source of generative modeling ability, and there are many alternative realizations in the deterministic way to regularize the loss with data distribution fitting error as that in AAE~\cite{makhzani2015adversarial} or ITL-AE~\cite{ITL_VAE2016}.
    By doing so, one can obtain deterministic or semi-deterministic autoencoders surmounting limitations of the VAEs but preserving their generative capability.
    
    We extend the two variants defined in Section~\ref{VAE_intro} to realize their deterministic counterparts: MMD-DAE and SWT-DAE for later use, motivating by the idea in~\cite{ITL_VAE2016}.
	The only thing needed is to change the way of generating latent representations for the encoder.
    That is, make the encoder output the latent directly instead of its mean and standard deviation $[\mu, \sigma]$ for the Gaussian PDF of the posterior distribution but train them with the same objective as the corresponding VAE does.

\subsection{Adversarial Robustness of Generative Autoencoders}
	Adversarial robustness is one of the key problems for neural networks. 
	It is common to generate adversarial examples by attacks to collapse VAEs or train them with adversarial examples to promote robustness.
    Notice that existing research on the adversarial robustness of generative autoencoders mostly focuses on the robustness of downstream classification with adversarial examples from the input space~\cite{kuzina2021diagnosing, ghosh2019resisting, sun2020type}.
    And studies involving the latent of an autoencoder aim to develop new methods for attack or defense by taking advantage of the latent regularization to obtain more adversarial-effect~\cite{gondim2018adversarial, kos2018adversarial,berthelot2018understanding}. 
    
    In this paper, we study the robustness of generative autoencoders directly from the latent space for the first time.
    Motivations come from real applications in communication systems where the issue of latent robustness or security arises as a problem.
    Because the information channel transmitting latent representations is exposed to noise interference or attackers as illustrated in Figure~\ref{Fig:Attack_scenarios}.
    We demonstrate that malicious latent can derail the decoder/generator of a generative autoencoder, and attempt to evoke research attention on their latent robustness. 

\section{Analysis for Robustness of VAEs in Latent space}\label{method: part-I}
	In this section, we show the vulnerability of the VAEs in the latent space through attack experiments at first.
	Then, a simple methodology is presented for adversarial training to improve the latent robustness. 
\subsection{Vulnerability of VAEs in the latent space}
\subsubsection{Problem Proposal}\label{sec_problem_definition}
	It is common to study the vulnerability of neural networks to adversarial samples through attack experiments. 
	Here we make the following assumptions and then define adversarial examples of the latent representations.
	\begin{itemize}
		\item{Assumption 1}. 
		One can get access to the latent (codes) of the encoder-decoder model.\label{Asmp1}
		\item{Assumption 2}. 
		One can get access to the decoder (needn't know the structure or functional in detail but can get the output of the decoder whenever given a specific latent code).\label{Asmp2}
	\end{itemize}
	
	As defined before, ${f_{enc}}(\cdot)$ and $f_{dec}(\cdot)$ denote the encoder and decoder of a well-trained VAE model, respectively. An un-targeted adversarial latent $z^{adv}$ to the original $z^{0}$ is defined as below:
	\begin{equation}\label{Eq.un_targeted_adv_latent_def}
	\left\{ 
	\begin{array}{l}
	J\left( {{z^{adv}},{z^0}} \right) = D\left( {{f_{dec}}\left( {{z^{adv}}} \right),{f_{dec}}\left( {{z^0}} \right)} \right)\\
	{z^*} = \mathop {\arg \max }\limits_{d\left( {z,{z^0}} \right)\leq\varepsilon} J\left( {z,{z^0}} \right)
	\end{array} 
	\right.,
	\end{equation}
	where $z^0=f_{enc}(x)$ can be encoded from an input $x$ or directly sampled from its prior distribution, and $D$ is some a distance or similarity measurement. 
	Any distance, divergence metrics for a two-sample test, or a composition of them can be used to realize $D$. 
	Without loss of generality, we use the mean square error in the experiments of this section, maintaining consistency with the reconstruction error term of the ELBO for the original VAE model. 
	One can easily change it for other metrics as needed. 
	Here $d$ measures the Euclidean distance between adversarial and the original latent, and $\varepsilon$ is a small positive number that represents a constraint on the attack intensity.
	The objective of a targeted adversarial latent is defined similarly:
	\begin{equation}\label{Eq.targeted_adv_latent_def}
	\left\{ 
	\begin{array}{l}
	J\left( {{z^{adv}}} \right) = D\left( {{f_{dec}}\left( {{z^{adv}}} \right),{x^t}} \right)\\
	{z^*} = \mathop {\arg \min }\limits_{d\left( {z,{z^0}} \right) < \varepsilon } J\left( z \right)
	\end{array} 
	\right.
	\end{equation}
	No matter whether planning to attack the decoder with a targeted example or in an un-targeted way, one should know the correct reconstruction as a prior. 
	That's why we need the Assumption~2.
	Note that the objective of the attack is to collapse the decoder, which means any alternatives that deteriorate the reconstructions or make a batch of generated samples being homogeneous are effective adversarial latent $z^{adv}$. 
	From this perspective, we can design kinds of targets for attacks such as images with all black/white pixels, with pixels randomly sampled from a prior distribution, or with the reversed color of the original reconstruction, and so on.
	In this section, we conduct two attack experiments to examine the latent robustness of the investigated VAE models. One is the un-targeted attack and the other is a targeted attack with all-black pixeled targets.
	
\subsubsection{Attack Experiment}\label{sec_Attack_experiments}
    We will take experiments on three well-trained models, which are the Vanilla-VAE 
    and the aforementioned MMD-VAE and SWT-VAE.

\textbf{Attack Method.}
	We choose PGD~\cite{madry2018towards} to optimize and solve Eq.(\ref{Eq.un_targeted_adv_latent_def}) $\sim$ Eq.(\ref{Eq.targeted_adv_latent_def}) for all the attack experiments in this work. 
    \begin{equation}\label{eq:PGD}
    	\left\{ \begin{array}{l}
    		z_0^{adv} = {z^0},\;\\
    		z_{k + 1}^{adv} = \mathop {Clip}\limits_{Z,\;\varepsilon } \left\{ {z_k^{adv} + \alpha sign\left( {{\nabla _Z}J\left( {Z_k,Z_{k-1}^{adv}} \right)} \right)} \right\}
    	\end{array} \right.
    \end{equation}
    Unless otherwise specified, parameter $\alpha$ is set to $\alpha=1$, the maximum iteration times is $k$=10, and $\varepsilon$ will be used to control the attack intensity (energy) for all scenarios. 

\textbf{Metrics to Evaluate Robustness.}
    When dealing with image issues, a good choice for robustness evaluation should related to the quality of the reconstructions or generations. 
    Candidates for such image quality evaluation can be PSNR~\cite{hore2010image}, SSIM~\cite{wang2004image}, IS~\cite{IS_salimans2016improved}, FID~\cite{FID_heusel2017gans}, LPIPS~\cite{zhang2018unreasonable} and so on. 
    In view of their excellent and reliable performance, we mainly take SSIM and LPIPS to score the image reconstruction quality in the current and future experiments. 
    
    The SSIM performs quite better than the MSE and PSNR in discriminating structural content in images.
    While the LPIPS is more effective to account for the nuances of human perception. 
    It is implemented with an ImageNet-trained deep neural network, e.g. VGG, but can also be used for other image datasets. 
    In this work, we use the realization from the TorchMetrics package~\cite{TorchMetrics_Measuring_2022} for Pytorch to calculate all the above-mentioned metrics.
   
    From a qualitative point of view, it is the truth that a VAE model is lack of adversarial robustness if the quality of the reconstructions decreases as the attack intensity increases, i.e., the SSIM score decreases or the LPIPS score increases. 
    Such curves can be viewed as distortion-to-distortion plots (DD-plots or DD-curves) like those in \cite{kos2018adversarial,gondim2018adversarial}. 
    However, quantitative evaluation of the latent robustness of the VAEs remains an open problem. In~\cite{gondim2018adversarial}, they address this problem with the AUDDC (Area under Distortion–Distortion Curve, AADDC). Motivated by this, we suggest the area associated with the DD-curves (AADDC) to quantify the latent robustness. For SSIM curves that achieve better performance with larger scores, our AADDC has the same definition as the AUDDC. While for LPIPS curves that achieve better performance with smaller scores, the AADDC denotes the area above the DD-curves.

\textbf{Attack Results Showing Vulnerability.}	
	We investigate the adversarial latent robustness of models trained on MNIST, FasionMNIST, and CelebA datasets. 
	All the models for each dataset share the same framework as shown in Figure~\ref{Fig:vae_framework}. 
	The encoder contains 4 \emph{convolutional} layers with hidden nodes of 32,64,128 and 64, respectively. 
	Each layer is followed by a \emph{BatchNorm2d} and a \emph{LeakyReLu} activation layer.
	The decoder is just the inverse of the encoder. 
	The only difference is the dimension of the latent, which is set to 10, 30, and 128 for models on MNIST, FasionMNIST, and CelebA, respectively.
	The batch size is set to 64 for training and is limited to 8 for attack experiments.
	The optimizer is Adam with learning rate $1e^{-3}$.

\begin{figure*}[tbp]
		\subfloat[]{\centering\includegraphics[width=0.3\linewidth]{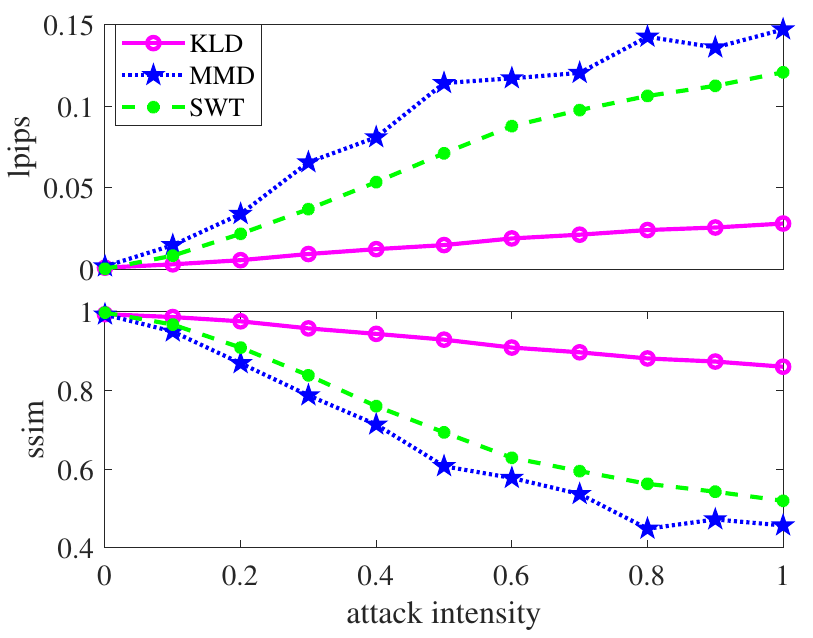} \label{Fig:vae-untargeted-mnist-ddplot}}
		\hfill
		\subfloat[]{\centering\includegraphics[width=0.3\linewidth]{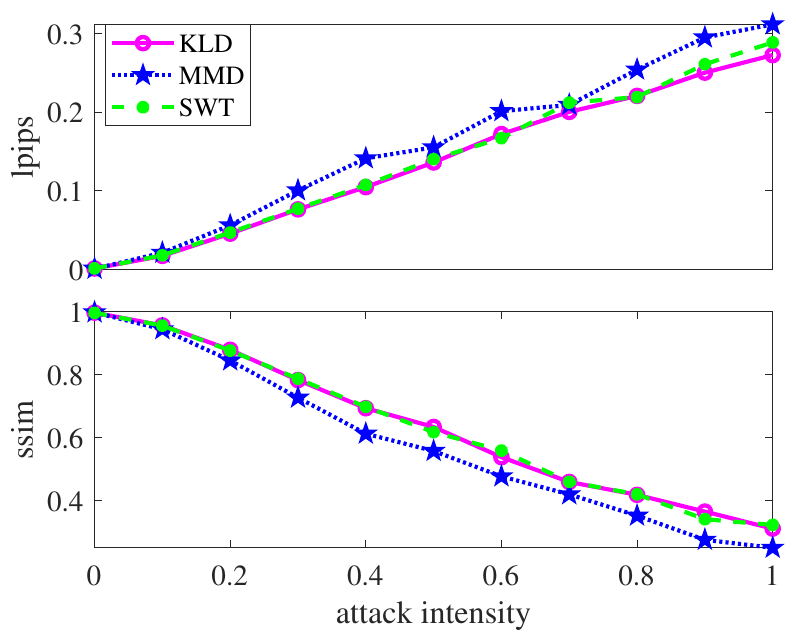} \label{Fig:vae-untargeted-fmnist-ddplot}}
		\hfill
		\subfloat[]{\centering\includegraphics[width=0.3\linewidth]{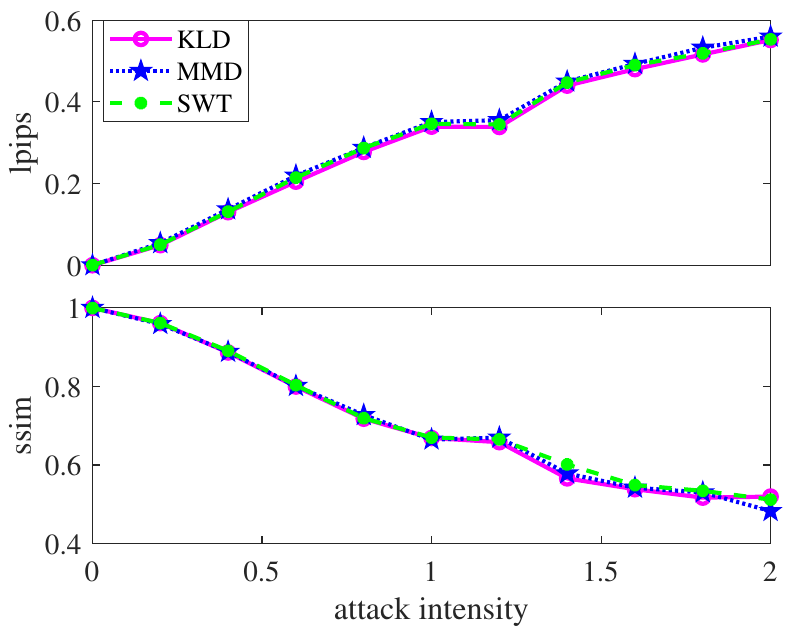} \label{Fig:vae-untargeted-celeba-ddplot}}
	
	    \vfill
		\subfloat[]{\includegraphics[width=0.3\linewidth]{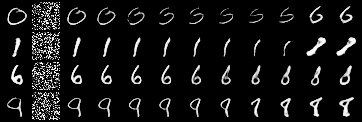} \label{Fig:vae-kld-untargeted-mnist-pairs}}
		\hfill
		\subfloat[]{\includegraphics[width=0.3\linewidth]{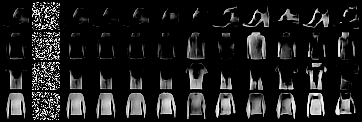} \label{Fig:vae-kld-untargeted-fmnist-pairs}}
		\hfill
		\subfloat[]{\includegraphics[width=0.3\linewidth]{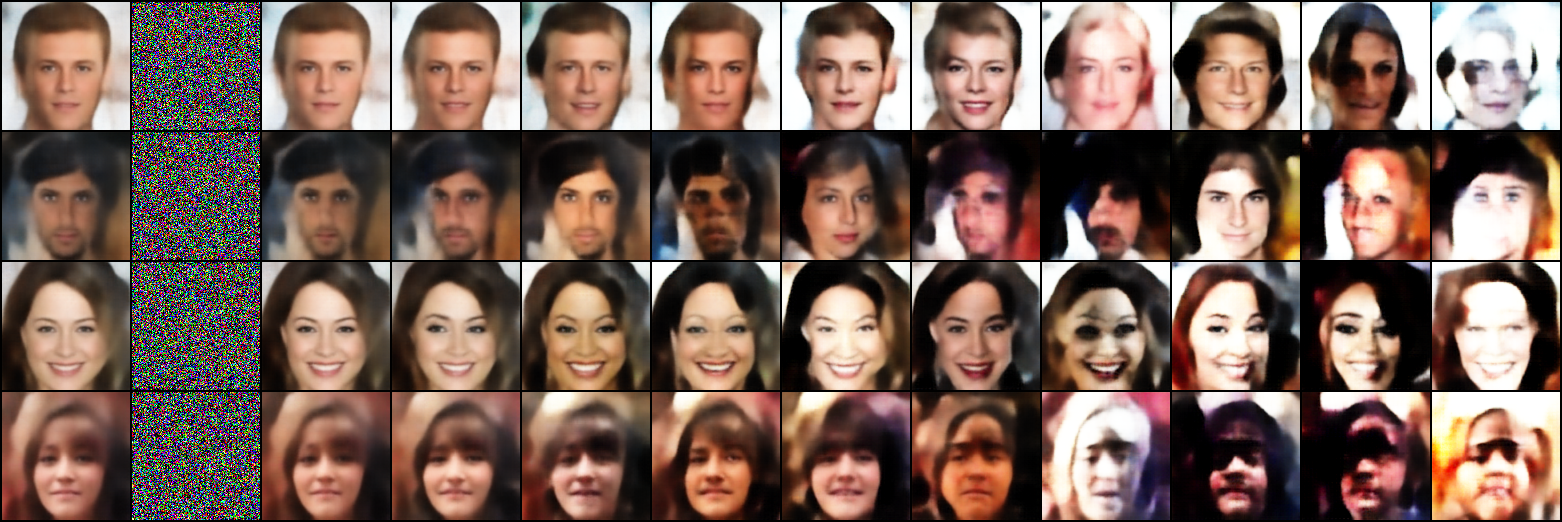} \label{Fig:vae-kld-untargeted-celeba-pairs}}
	    \vfill
	
		\subfloat[]{\includegraphics[width=0.3\linewidth]{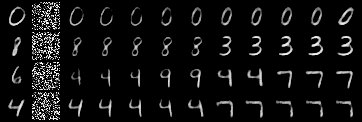} \label{Fig:vae-mmd-untargeted-mnist-pairs}}
		\hfill
		\subfloat[]{\includegraphics[width=0.3\linewidth]{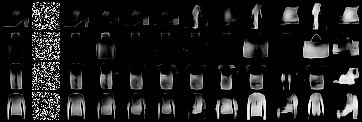} \label{Fig:vae-mmd-untargeted-fmnist-pairs}}
		\hfill
		\subfloat[]{\includegraphics[width=0.3\linewidth]{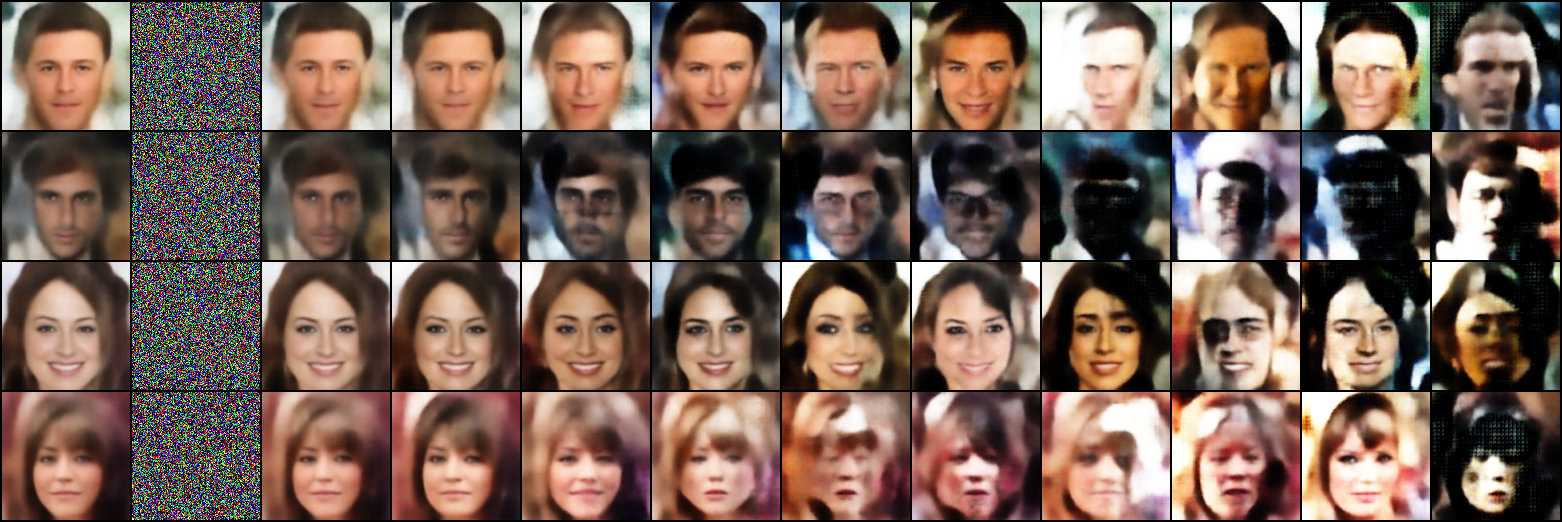} \label{Fig:vae-mmd-untargeted-celeba-pairs}}
	    \vfill
		\subfloat[]{\includegraphics[width=0.3\linewidth]{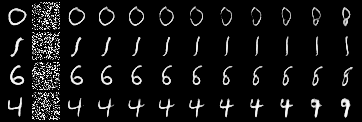}	\label{Fig:vae-swt-untargeted-mnist-pairs}}
		\hfill
		\subfloat[]{\includegraphics[width=0.3\linewidth]{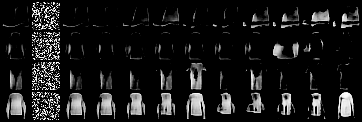} \label{Fig:vae-swt-untargeted-fmnist-pairs}}
		\hfill
		\subfloat[]{\includegraphics[width=0.3\linewidth]{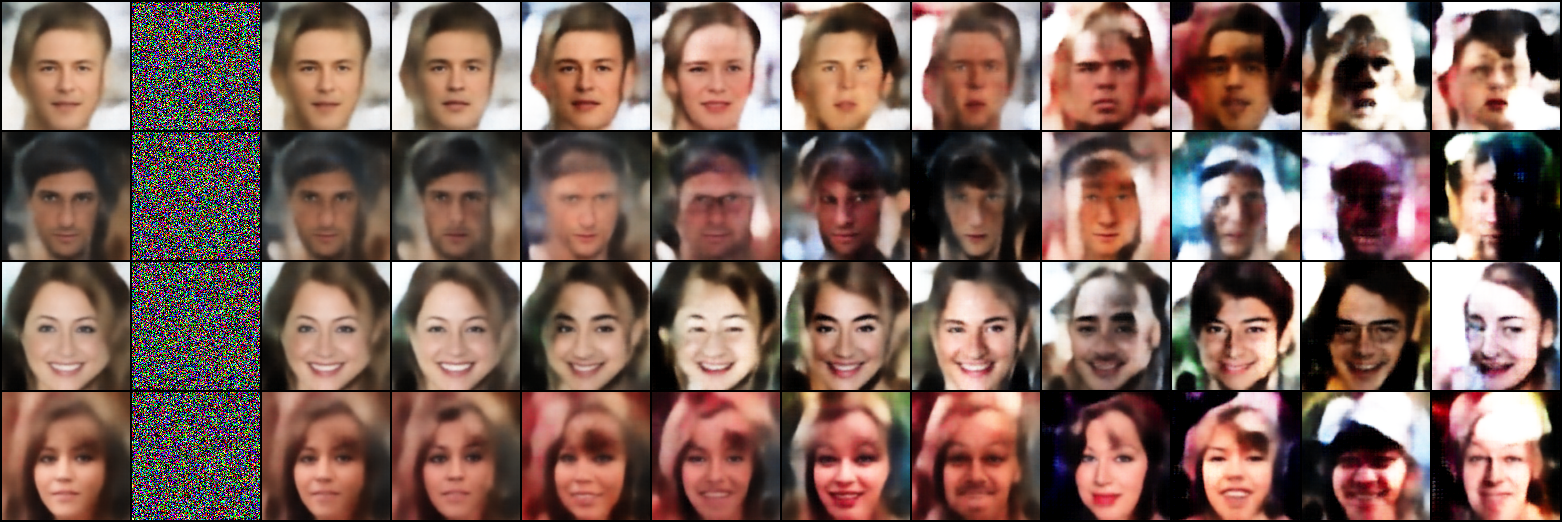} \label{Fig:vae-swt-untargeted-celeba-pairs}}


	\caption{Results of the un-targeted attack experiments. 
			(a), (b), (c) are DD-plots for the investigated three VAEs trained on MNIST, FasionMNIST, and CelebA dataset, respectively.
			Attack intensity for the the first two datasets are $\varepsilon=[0,1]$ with step=0.1, and $\varepsilon=[0,2]$ with step=0.2 for the CelebA. 
			A small LPIPS or big SSIM score implies good reconstruction quality. 
			(d), (e), (f) Show adversarial reconstruction pairs of one single batch from KLD-VAE correspond to \ref{Fig:vae-untargeted-mnist-ddplot}, \ref{Fig:vae-untargeted-fmnist-ddplot} and \ref{Fig:vae-untargeted-celeba-ddplot}. 
			Images in the first columns are original reconstructions, and adversarial images of different attack intensity are displayed from the third columns to the last. In the second column is the images with random noised to separate the original and adversarial reconstructions.
			(g), (h), (i) Show adversarial reconstruction pairs of one single batch from different datasets on MMD-VAE. 
			(j), (k), (l) Show adversarial reconstruction pairs from SWT-VAE. 
			}
        \label{Fig:adv-rec-pairs}
	\end{figure*}

	At first, we take an experiment to attack the well-trained models in an un-targeted way. 
	As shown in Figure~\ref{Fig:adv-rec-pairs},	it exhibits quality deterioration and significantly different reconstructions when the attack goes strong on all three datasets. 
	Figure~\ref{Fig:vae-untargeted-mnist-ddplot},~\ref{Fig:vae-untargeted-fmnist-ddplot},~\ref{Fig:vae-untargeted-celeba-ddplot} directly support the judgment that adversarial reconstructions show a trend of increasing difference from the original ones as the attack intensity increases. 
	But the three VAE models regularized with different terms perform quite differently. 
	It seems that the reconstruction-quality scores are getting more and more similar among different models as the complexity of the data increases.
	For instance, the curves for different models on MNIST in Figure~\ref{Fig:vae-untargeted-mnist-ddplot} can be recognized with a clear distinction while they perform very close on FMNIST as in Figure~\ref{Fig:vae-untargeted-fmnist-ddplot}, and in Figure~\ref{Fig:vae-untargeted-celeba-ddplot} for CelebA they even overlaps. 
	Furthermore, as shown in the last three rows of Figure~\ref{Fig:adv-rec-pairs}, reconstructions from different types of models based on the three datasets are not only differed from the original images but also qualitatively declined.
	
\begin{figure*}[tbp]
		\centering
	    \begin{minipage}{0.90\linewidth}
		\centering
		\subfloat[]{\centering\includegraphics[width=0.4\linewidth]{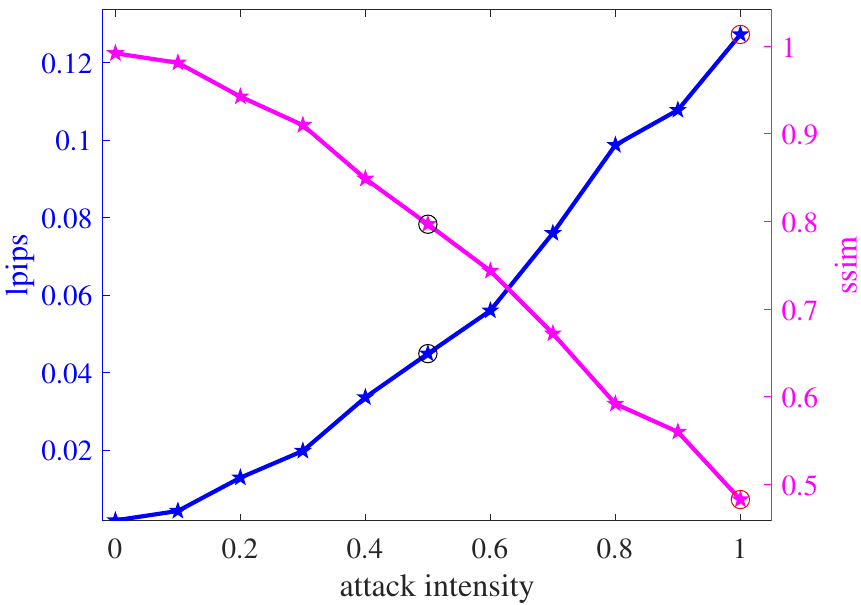} \label{Fig:vae-kld-mnist-black-ddplot}}
		\hfill
		\subfloat[]{\centering\includegraphics[width=0.5\linewidth]{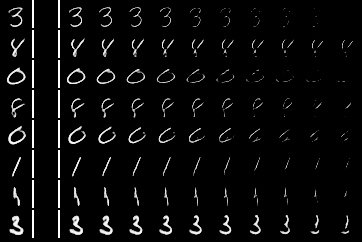} \label{Fig:vae-kld-mnist-black-adv_pairs}}


        \end{minipage}
		\caption{Results of the targeted-attack experiments on a well-trained KLD-VAE with MNIST dataset. 
				(a) DD-plots based on fidelity scores of the adversarial reconstructions with all-black targets.
				(b) Some of the adversarial reconstruction pairs under ten attack experiments. The first column displays original reconstructions, in the second column are the attack targets splitted by two white lines from other images, and the rest columns are adversarial reconstructions under attack of different intensity with $\varepsilon\in\left[0, 1\right]$ with step $0.1$.
				}
		\label{Fig:vae-kld-mnist}
	\end{figure*}

	The above experiment has proved that un-targeted attacks in the latent space are effective to fail the reconstruction or generation of VAE models.
	Next, we investigate the latent robustness of VAEs to the targeted attack.
	Figure~\ref{Fig:vae-kld-mnist} presents the results of adversarial reconstructions and the corresponding DD-plots of the Vanilla-VAE under attack with all-black targets on the MNIST dataset. 
	It can be concluded that the VAEs are indeed prone to be attacked in the latent space.
	The LPIPS and SSIM scores in Figure~\ref{Fig:vae-kld-mnist-black-ddplot} show an explicit worsen trend as attack intensity arises.
	As displayed in Figure~\ref{Fig:vae-kld-mnist-black-adv_pairs}, model reconstructions under attacks is deteriorating, too.
    And the investigated model reconstructs images with nearly all black pixels under a black-targeted attack when the intensity rises to $\varepsilon=1.0$. Aware that this adversarial attack is added in the latent space, so its intensity no longer represents the scale of the image pixels in the input space. And we will give a visualization of the adversarial latent in the subsequent experiment.
    
    Effective and efficient targets for attack are always deep associated with the features of the dataset. 
    Here we have demonstrated that all-black targets are of this category for the MNIST dataset.
    Since the aim for us is to verify the deceptive reconstruction capability with adversarial latent generated by the targeted attack and this has been achieved, we will show no more results with other types of targets for MNIST and the other two datasets. 
    It worth believed that targeted attacks are definitely capable of collapse the reconstruction or generation of VAEs as long as the right target and enough attack power is taken.  

\begin{figure*}[tbp]
		\centering
		\begin{minipage}{0.60\linewidth}
		\includegraphics[width=1.0\linewidth]{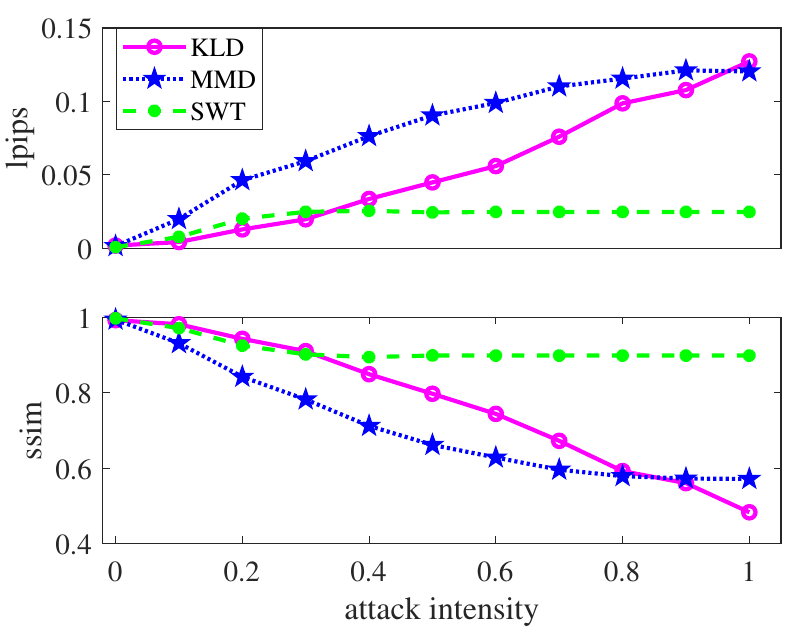}
		\end{minipage}
		\caption{Fidelity scores of reconstruction from well-trained VAEs on MNIST dataset under all-black targeted-attack. The Vanilla-VAE, MMD-VAE, and SWT-VAE all face a performance decline with the increasing of the attack intensity.}
		\label{Fig:attack-DDplots}
	\end{figure*}

	The DD-plots based on SSIM and LPIPS scores in Figure~\ref{Fig:vae-kld-mnist-black-ddplot} qualitatively demonstrate the vulnerability of VAE models in the latent space. 
	However, an quantitative metric of the latent robustness is usually necessary when evaluating the difference among several models. 
	For instance, the image quality score statistics for three different VAE models under attack are illustrated in  Figure~\ref{Fig:vae-untargeted-mnist-ddplot},~\ref{Fig:vae-untargeted-fmnist-ddplot},~\ref{Fig:vae-untargeted-celeba-ddplot} and Figure~\ref{Fig:attack-DDplots}, how can we tell the quantitative difference among them? 
	This is just the motivation for us to propose the statistical AADDC. 

    As presented in Table~\ref{tab:AADDC-VAE}, we compute the AADDC scores for each model under attack with all black targets on three datasets, with the help of which we can provide a quantitative evaluation on the adversarial latent robustness of the investigated models. 
    If judge them from the latent robust perspective, we get an intuitive conclusion that SWT-VAE performs the best on the MNIST dataset, KLD-VAE the second and MMD-VAE the worst. 

	\small
        {\begin{table}[tbp]
	    \centering
	    \caption{Statistics of AADDC for the three models trained on three different datasets under attacks with all black targets. Limit lines are set to be LPIPS=0.15 and SSIM=0.4 as shown in Figure~\ref{Fig:AADDC-example}. }
		\label{tab:AADDC-VAE}
		\begin{tabular}[width=0.99\columnwidth]{@{}ccccccc@{}}
			\toprule
			\multirow{2}{*}{} & \multicolumn{2}{c}{MNIST} & \multicolumn{2}{c}{FasionMNIST} & \multicolumn{2}{c}{CelebA} \\ 
			                  & LPIPS       & SSIM        & LPIPS          & SSIM           & LPIPS        & SSIM        \\ \cmidrule(r){1-7}
			KLD-VAE           & 0.098       & 0.378       & 0.098          & 0.259          & 0.134        & 0.433       \\
			MMD-VAE           & 0.069       & 0.305       & 0.115          & 0.260          & 0.115        & 0.438       \\
			SWT-VAE           & 0.129       & 0.514       & 0.098          & 0.262          & 0.122        & 0.435       \\ 
			\bottomrule	
		\end{tabular}
	\end{table} } 

\begin{figure}[tbp]
		\centering
		\begin{minipage}{0.60\linewidth}
		\includegraphics[width=1.0\columnwidth]{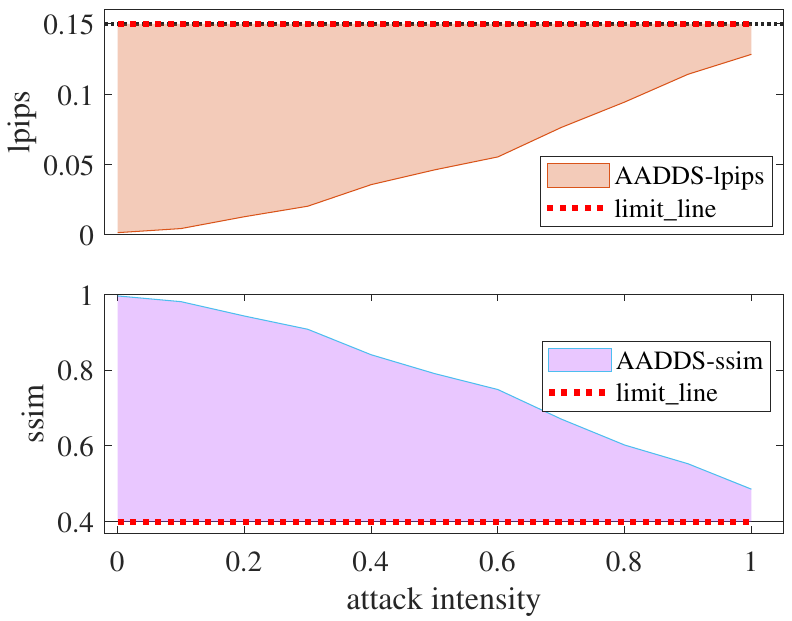}
		\end{minipage}
		\caption{An example of LPIPS and SSIM based AADDC visualization and the corresponding integration limit line, LPIPS=0.15 and SSIM=0.4.}
		\label{Fig:AADDC-example}
	\end{figure}

    When calculating the statistical AADDC for the LPIPS curve, a base line (the red dotted) is required to set above them as plotted in the upper sub-figure of Figure~\ref{Fig:AADDC-example}, and similarly a base line under the SSIM curve is set in the lower sub-figure. 
    The associated AADDC scores are then obtained by integrating the area between the measurement curve and the base line.

\begin{figure*}[btp]
		\centering
		\begin{minipage}{0.90\linewidth}
    		\subfloat[all-black targeted attack]{\centering\includegraphics[width=0.48\linewidth]{ 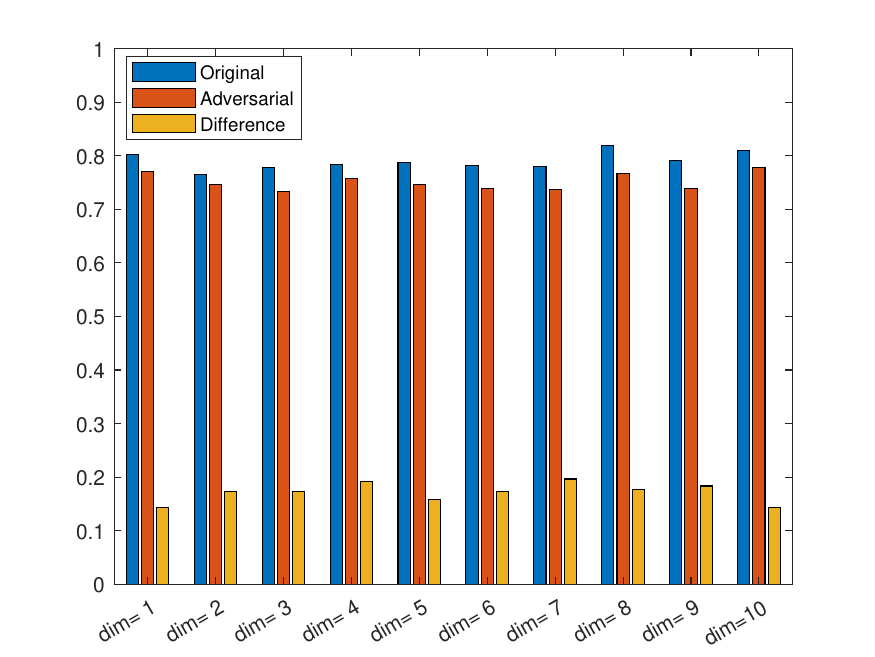 }\label{Fig:latent-visualize-kld-black-mnist-0.5}}
    		\hfill
    		\subfloat[un-targeted attack]{\centering\includegraphics[width=0.48\linewidth]{ 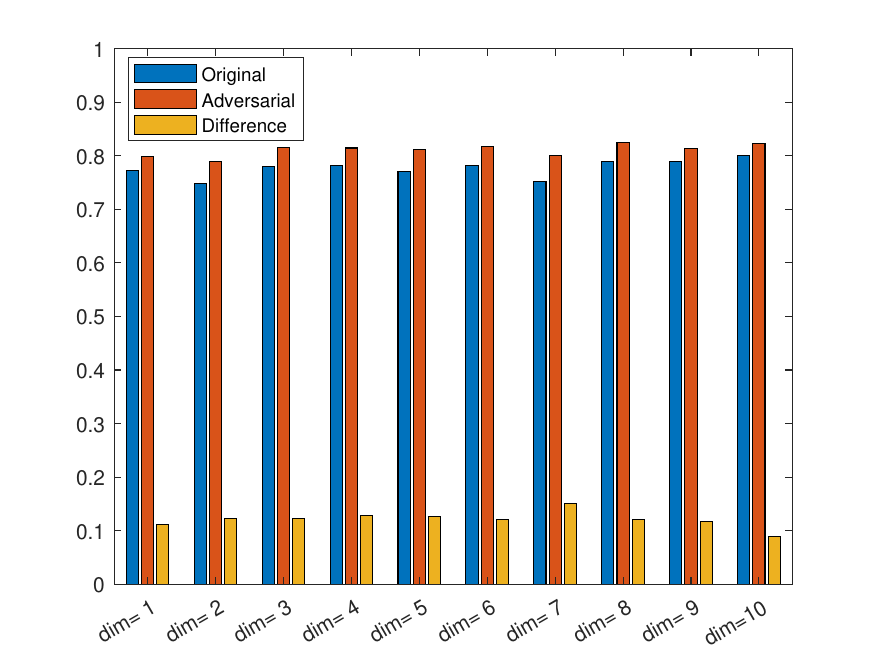 }\label{Fig:latent-visualize-kld-untarget-mnist-0.5}}
		\end{minipage}
		\caption{Dimension-wise mean absolute latent statistical visualization of three types of attack on the Vanilla-VAE trained on the MNIST dataset with attack intensity $\varepsilon=0.5$.}
		\label{Fig:latent-visualize-kld-mnist-0.5}
	\end{figure*}

\begin{figure*}[btp]
		\centering
		\begin{minipage}{0.90\linewidth}
    		\subfloat[all-black targeted attack]{\centering\includegraphics[width=0.45\linewidth]{  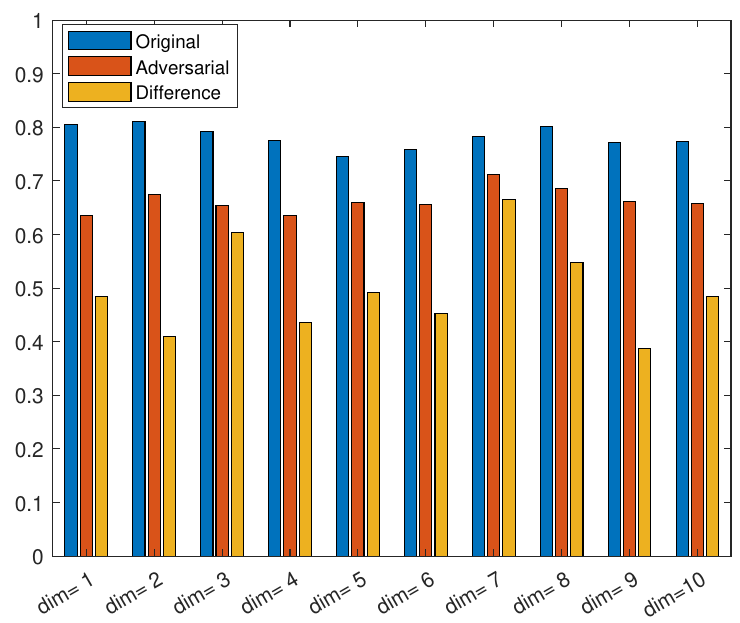}
    		\label{Fig:latent-visualize-kld-black-mnist}}
    		\hfill
    		\subfloat[un-targeted attack]{\centering\includegraphics[width=0.45\linewidth]{ 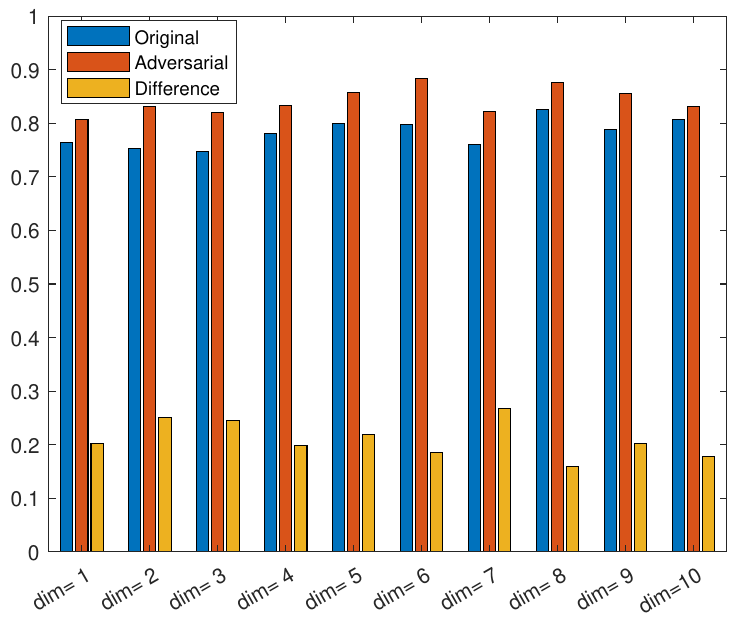}\label{Fig:latent-visualize-kld-untarget-mnist}}
		\end{minipage}
		\caption{Dimension-wise mean absolute latent statistical visualization of three types of attack on the Vanilla-VAE trained on the MNIST dataset with attack intensity $\varepsilon=1.0$ . }
		\label{Fig:latent-visualize-kld-mnist}
	\end{figure*}

\textbf{Visualization of Latent Intervention.}
	A straightforward question arises what the adversarial latent looks like and how far away it is from the original one. 
    Unlike adversarial examples for input images, adversarial latent cannot be visualized in an intuitive way. 
    Consider a mini-batch of latent codes $Z \in R^{m \times d}$, where $m$ is the batch size and $d$ is the dimension of the latent, and then we compute the dimension-wise average latent absolute as below:
	\begin{equation}
	\left\{ \begin{array}{l}
		E(\left| {{Z_j}} \right|) = \frac{1}{m}\sum\limits_{i = 1}^m {\left| {{z_{ij}}} \right|} \;\;\\
		E(\left| {Z_j^{adv}} \right|) = \frac{1}{m}\sum\limits_{i = 1}^m {\left| {z_{ij}^{adv}} \right|} \;\;\\
		E(\left| {{\delta _j}} \right|) = \frac{1}{m}\sum\limits_{i = 1}^m {\left| {z_{ij}^{adv} - {z_{ij}}} \right|} 
		\end{array} \right.,j = 1,2, \cdots ,d.
	\end{equation}
	With this mean-absolute difference visualized in Figure~\ref{Fig:latent-visualize-kld-mnist-0.5}$\sim$Figure~\ref{Fig:latent-visualize-kld-mnist}, it is shown that an effective attack can be realized with small modification in latent representation.
    The mean absolute variations are tiny as shown in Figure~\ref{Fig:latent-visualize-kld-mnist-0.5} at attack intensity $\epsilon=0.5$ while the effectiveness of attacks at this intensity is still significant as shown in Figure~\ref{Fig:adv-rec-pairs} and Figure~\ref{Fig:vae-kld-mnist}.
	Although the mean absolute latent difference between the adversarial and original are significant at intensity $\varepsilon=1.0$ for all-black targeted attack on the Vanilla-VAE with the MNIST dataset, we have obtained adversarial reconstructions with almost all-black pixels as shown in the bottom row of Figure~\ref{Fig:vae-kld-mnist-black-adv_pairs}. 
	Aware that the all-black pixeled target may be too harsh a choice for attack on the MNIST dataset.
	
\begin{figure*}[btp]
		\centering
		\subfloat[On FasionMNIST]{
					\centering\includegraphics[width=0.45\linewidth]{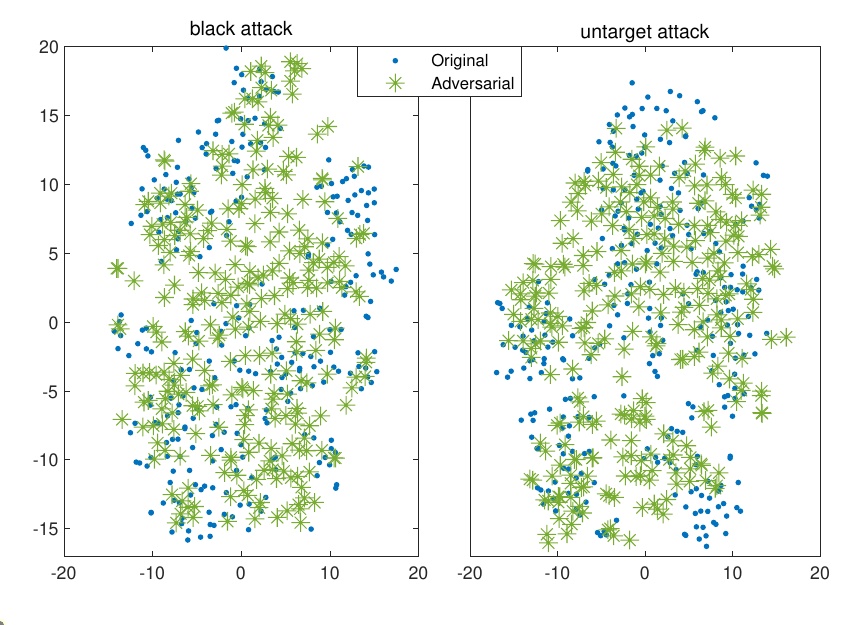 }
					\label{Fig:latent-visualize-kld-fmnist}
					}
		\hfill
		\subfloat[On CelebA]{
					\centering\includegraphics[width=0.45\linewidth]{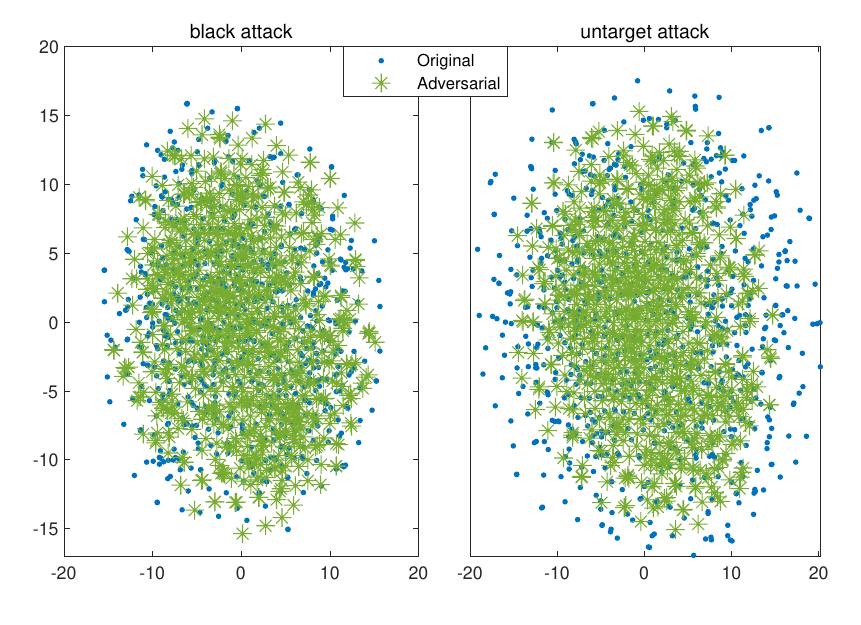 }
					\label{Fig:latent-visualize-kld-clceba}
					}
		\caption{Latent visualization of the Vanilla-VAE trained on the FasionMNIST and CelebA under three types of attack with intensity $\varepsilon=1.0$.}
		\label{Fig:latent-visualize-kld-fmnist+celeba}
	\end{figure*}

	It's obvious that the dimension-wise visualization of the latent is impractical when the dimensionality is too high. 
	Thus, we use the t-SNE method~\cite{van2008visualizing} to map the latent representations to two-dimensional variables for the FMNIST and CelebA dataset, and then visualize them in Figure~\ref{Fig:latent-visualize-kld-fmnist+celeba}. 
	Overall, the following phenomenon can be identified from the figures: 
	(a) The difference between adversarial and the original latent is tiny on FMNIST dataset under both types of attacks.
	(b) The scattered range of the adversarial latent for both datasets seems to be smaller than the original, which may imply that the attacks make the generated/reconstructed samples less diverse. 

\subsection{Latent Adversarial Training Autoencoder}
\begin{figure*}[tbp]
		\centering
	    \begin{minipage}{0.99\linewidth}
		\centering
		\subfloat[LPIPS]{\centering\includegraphics[width=0.5\linewidth]{ 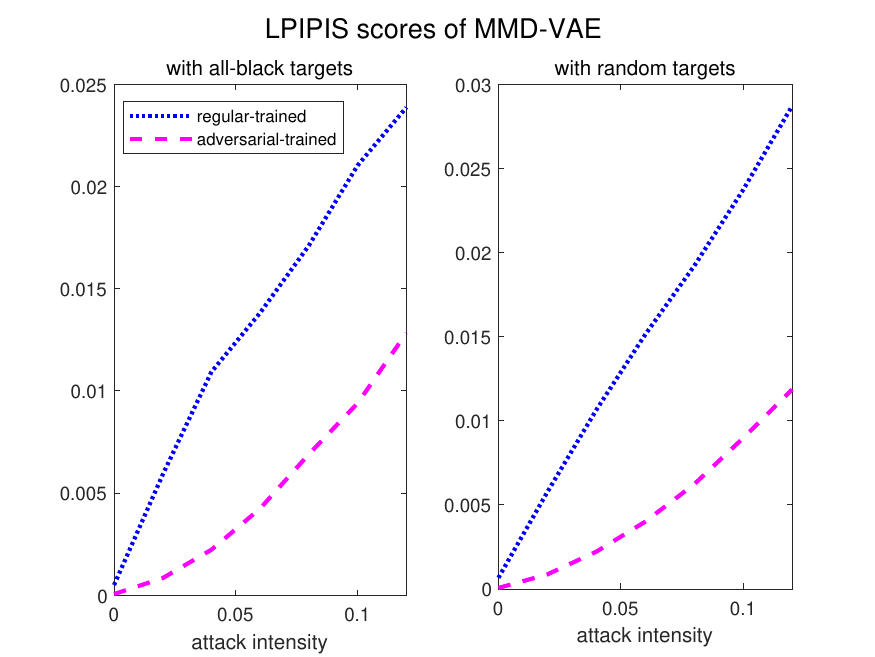}\label{Fig:LPIPS-mmd-adv-trained}} 
		\hfill
		\subfloat[SSIM]{\centering\includegraphics[width=0.5\linewidth]{ 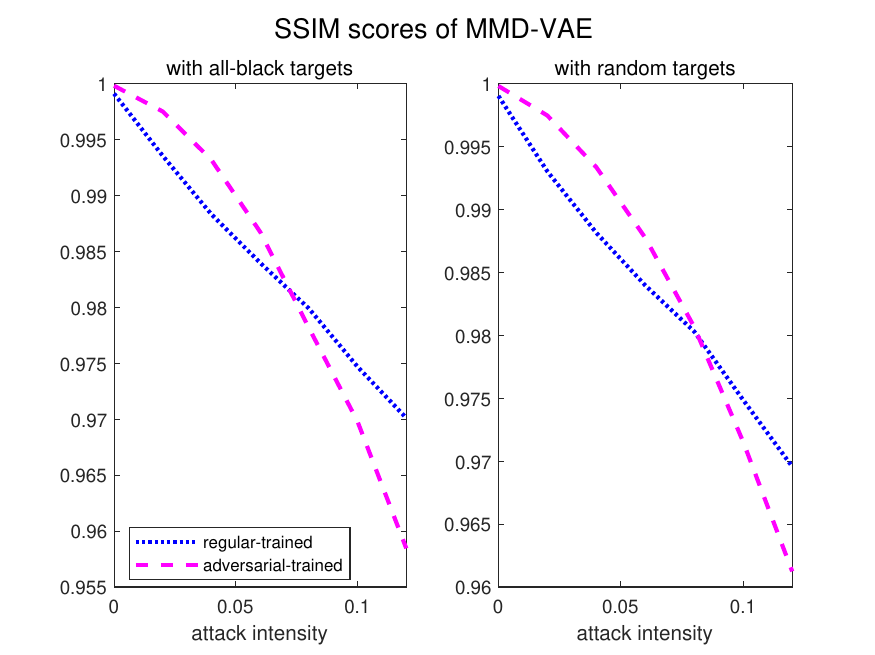}\label{Fig:ssim-mmd-adv-trained}}
		\caption{Fidelity scores of regular and adversarial-trained VAEs under targeted attack. The similarity Metric used in PGD attack is SSIM.}
		\label{Fig:adv-cmp}
        \end{minipage}
	\end{figure*}

	In this section, we show the ability of latent robustness promotion by adversarial training. 
	Consider adding an adversarial training loop to a VAE model.
	After the regular training procedure at each mini-batch, the decoder is optimized for given iteration times with the loss of adversarial reconstructions from original samples, thus achieving more robustness.
	The pseudo-code is shown as in Algorithm~\ref{alg:LAT-VAE}. 
	We use the PGD attack to generate adversarial latent codes in an un-targeted way, and the SSIM is taken to measure the distance, $D$ as defined in Eq~(\ref{eq:PGD}), between original reconstructions and adversarial ones. 
\small{
    \begin{algorithm}[tbp]
    	\caption{\;\;Latent Adversarial Training Autoencoder}\label{alg:LAT-VAE}
    	\textbf{Require:}  $X$=training data, $N$=number of batches. $m$=size of the mini-batch. $M$=maximum adversarial training iterations. $K$=maximum iteration numbers for the PGD Attack optimization. $e$ denotes the reconstruct loss measure metric MSE in this work, $r$ is the regularizer like KLD, MMD or SWT, and $\lambda$ is the weight for regularization. $\theta$, $\varphi$ are the network parameters for $f_{enc}$ and $f_{dec}$, respectively.     \\   
    	\textbf{Initialize:}  $\theta, \varphi$ and hyper-parameters $m, M, K, \lambda, \alpha, \varepsilon$.         
    	\begin{algorithmic}[1]
    		\For {$i=0,\cdots,N$}\\ \;\;\;\;\;
    			Sample next mini-batch $X_m$ from $X$ \\ \;\;\;\;\;
    			${Z_m} = {f_{enc}}\left( {{X_m}} \right)$, \;\;${{\hat X}_m} = {f_{dec}}\left( {{Z_m}} \right)$ \\ \;\;\;\;\;
    			$\left( {\theta ,\varphi } \right) \leftarrow Adam\left( {{\nabla _{\left( {\theta ,\varphi } \right)}}\left[ {e\left( {{X_m},{{\hat X}_m}} \right) + \lambda r\left( {{Z_m}\left| {{X_m}} \right.} \right)} \right]} \right)$ 
    			\For {$j=0,\cdots,M$} \\ \;\;\;\;\;\;\;\;\;\;
    			    $Z_0^{adv} = {Z_m}$ updated by re-parameterize process with the encoded $[\mu, \sigma]$ of ${{Z_m}\left| {{X_m}} \right.}$.	
        			\For {$k=0,\cdots,K$}		\\
        				\;\;\;\;\;\;\;\;\;\;\;\;\;\;\;$Z_{k + 1}^{adv} = \mathop {Clip}\limits_{Z,\;\varepsilon } \left\{ {Z_k^{adv} + \alpha sign\left( {{\nabla _Z}J\left( {Z_k,Z_{k-1}^{adv}} \right)} \right)} \right\}$ \\  \;\;\;\;\;\;\;\;\;\;\;\;\;\;
        				$\hat{X}_m^{adv} = f_{dec}(Z_K^{adv}) $
        			\EndFor \\ \;\;\;\;\;\;\;\;\;\;
        			$\varphi  \leftarrow Adam\left( {{\nabla _\varphi }e\left( {{{ X}_m},\hat{X}_m^{adv}} \right)} \right)$
    			\EndFor
    		\EndFor%
    	\end{algorithmic}			
    \end{algorithm}	
    }
    We conduct attack experiments to compare the robustness difference between the original-trained and adversarial-trained models. 
    The results are presented in Figure~\ref{Fig:adv-cmp}. 
    The LPIPS curves encourage us to believe that the model with adversarial training performs significantly more robust under attacks than the regular one. 
    But the SSIM curves of the adversarial-trained model only outperform the regular-trained one at the first phase and then give a worse performance as shown in Figure~\ref{Fig:ssim-mmd-adv-trained}. 
    This is because the model was trained with all adversarial latent generated under attacks with an intensity of $\varepsilon=0.05$. 
    As a result, the adversarial-trained model performs more robust under attacks with small or similar intensities but is powerless against attacks with too much bigger intensities. 
    
    Anyway, this experiment has shown the potential of promotion on latent robustness through adversarial training. 
    It can be expected that there are many other effective adversarial training methods to obtain latent robust VAE models, and we leave it for future work. 

\section{Deeper Analysis on Latent Robustness}	
\subsection{Comparison of Latent Robustness between VAEs and DAEs}\label{method: part-II}
	Conventional autoencoders are of deterministic structure, and the VAE pioneered a variational functional structure. 
	Thanks to this structure, VAEs not only achieve higher robustness to input space perturbations but are also able to generate new reasonable examples.
	Despite high expectations from the day it was developed, VAE is still very difficult to be used in practical applications.
	In recent years, many scholars have turned their attention back to autoencoders with deterministic structure~\cite{norouzi2020exemplar,polykovskiy2020deterministic,ghosh2020variational,wu2019couple,ding2021semi}. 
	Perhaps analyzing and understanding the difference between VAEs and DAEs from the perspective of latent robustness could give us new insights.
	
	Whether the VAEs or DAEs are more robust can be observed through attack experiments on them and then evaluate their latent robustness performance.
	To conduct a fair comparison, we will take experiments on two types of models that share the same framework except the way to generalize the latent codes, that is, the variational and deterministic ways, respectively.
    Consequently, the DAEs with the same structure as mentioned in section~\ref{section-DAE} can help to realize this plan.
	What we need to do is choose the same regularization in loss functions for the corresponding VAE and DAE model pairs.

\begin{figure*}[tbp]
		\centering
	    \begin{minipage}{0.99\linewidth}
	    \subfloat[MMD Regularized]{\centering\includegraphics[width=0.5\linewidth]{ 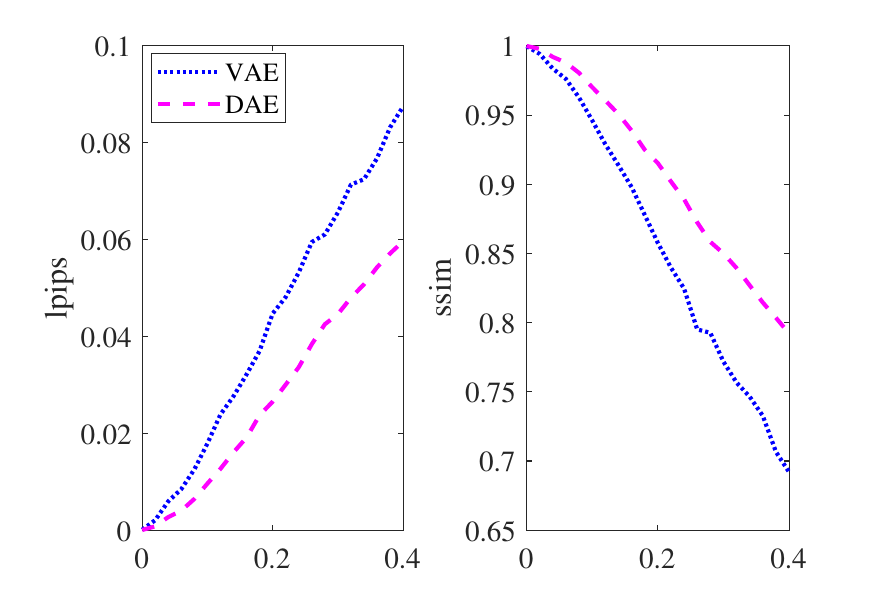 }\label{fig:vae-dae-mmd-black-mnist}}
	    \hfill
	    \subfloat[SWT Regularized]{\centering\includegraphics[width=0.5\linewidth]{ 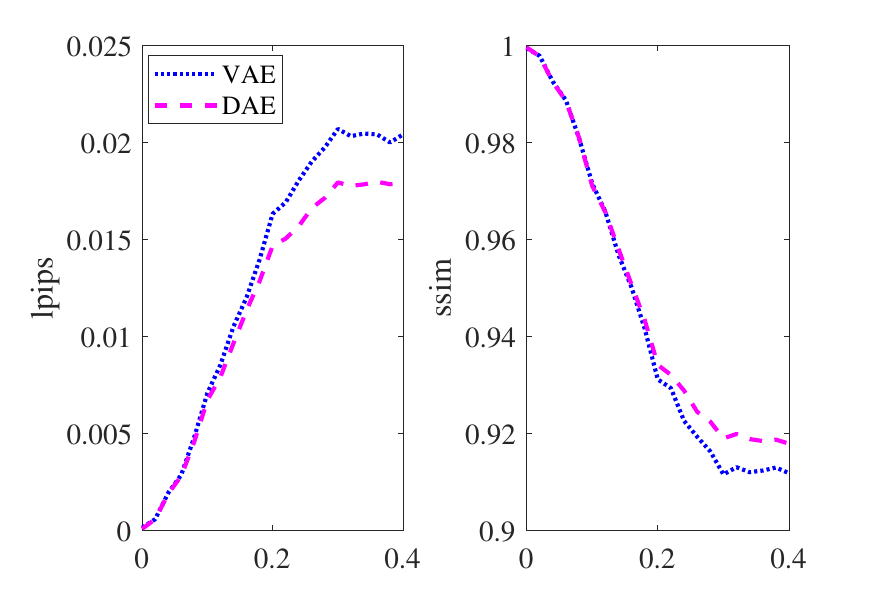 }\label{fig:vae-dae-swt-black-mnist}}
		\caption{Fidelity scores comparison between VAEs and DAEs under black-targeted attack.}
		\label{fig:vae-dae-mnist}
		\end{minipage}
	\end{figure*}

	Experiment settings for attack are the same as that in Section~\ref{method: part-I}, and the results under black targets on the MNIST dataset are depicted in Figure~\ref{fig:vae-dae-mnist}. 
	Both the LPIPS and SSIM-based DD-curves imply that deterministic autoencoders are more robust than variational autoencoders. 
	We also take another two attack experiments based on FasionMNIST and CelebA datasets. 
	The DD-curves are similar to those in Figure~\ref{fig:vae-dae-mnist}, so we do not plot them here but the statistical AADDC scores are presented in Table~\ref{tab:vae-dae}. 
	
	 All the statistics give us an insight that deterministic autoencoders perform more robustly in the latent space than variational autoencoders.
	 In this sense, our results further justify another potential advantage of DAE over VAE in terms of latent adversarial robustness.
	
	In addition, the robustness differences between VAEs and DAEs tend to decline as the increase of latent dimensionality in view of AADDC. 
	For instance, the dimensionality of $z$ for models on MNIST, FMNIST, and CelebA is set to be 10, 30, and 128, respectively, and the measure gap of the AADDC between VAEs and DAEs seems to be narrowed as it grows.
	This is quite similar to what happens in the robustness comparison of different VAEs in Figure~\ref{Fig:vae-untargeted-mnist-ddplot},~\ref{Fig:vae-untargeted-fmnist-ddplot},~\ref{Fig:vae-untargeted-celeba-ddplot}. 
	
    \small{
    \begin{table}[tbp]
    \setlength\tabcolsep{3pt}   
    \centering
    \caption{AADDC based on LPIPS and SSIM scores for attack experiments on three VAE models and their deterministic counterparts. When calculating the AADDC, limit-lines are LPIPS=0.15 and SSIM=0.4 consistent with the previous settings.}
    \label{tab:vae-dae}
    \begin{tabular}{@{}ccccc@{}}
        \toprule
        \multirow{2}{*}{VAE/DAE} & \multicolumn{2}{c}{LPIPS-AADDC} & \multicolumn{2}{c}{SSIM-AADDC} \\ 
                                 & MMD          & SWT           & MMD           & SWT       \\ \cmidrule(l){1-5}
        MNIST                    & 0.0431/0.0490      & 0.0432/0.0489   & 0.1830/0.2034   & 0.1833/0.2033           \\
        FMNIST                   & 0.0329/0.0339      & 0.0316/0.0334   & 0.1679/0.1697   & 0.1653/0.1659           \\
        CelebA                   & 0.0375/0.0376      & 0.0366/0.0370   & 0.2243/0.2252   & 0.2228/0.2231           \\ \bottomrule
    \end{tabular}
    \end{table}	
    }
\subsection{Relation between Latent Robustness and Disentanglement}\label{method: part-III}
\begin{figure*}[tbp]
	    \centering\includegraphics[width=0.9\linewidth]{ 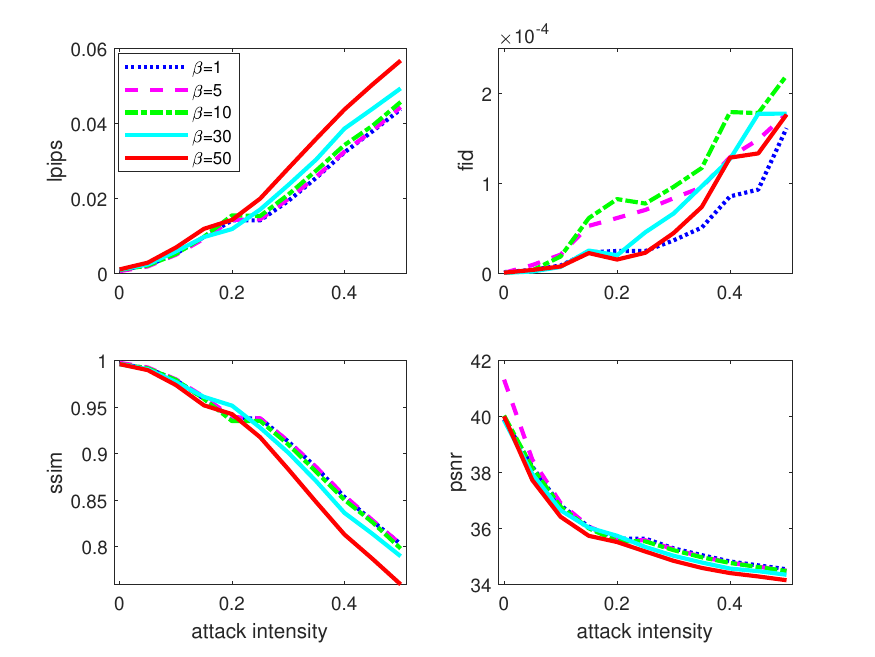 }
	    \caption{Fidelity scores of reconstruction from $\beta$-TC-VAEs under black targeted-attacks based on MNIST.}
	    \label{fig:disentangle-betatcvae}
	\end{figure*}

	Disentangled representation is one of the key pursuits for machine learning~\cite{bengio2013representation}.
	A model with disentangling ability can obtain semantic features from the original high-dimensional data. 
    It helps to achieve the interpretability of the representation network.
    One can also generate new samples or change existing ones toward a demanded style by manipulating some specific dimensional features of the semantic disentangled latent.
	It is a common sense that there is a trade-off between the reconstruction accuracy and disentanglement. 
	This implies that once we try to improve the disentangling effect of the encoded latent representations, the decoding reconstruction error increases. 
	Additionally, researchers have found that $\beta-$TCVAE with larger $\beta$ is more robust to adversarial input~\cite{willetts2020improving, willetts2019disentangling}. Are there trade-offs among the reconstruction loss, the disentanglement and the latent robustness?
	To answer the above question, we conduct experiments to attack the $\beta$-TCVAE~\cite{willetts2019disentangling}.
	As shown in Figure~\ref{fig:disentangle-betatcvae}, the LPIPS (SSIM) scores grow faster (slower) with a larger weight for the TC term $\beta$, which controls the disentangling strength. 
	Though not monotonic as the LPIPS and SSIM, the FID and PSNR scores show a similar trend. 
	Now, we may conclude that the disentangling strength of the latent representation in VAEs does damage the latent adversarial robustness. 
	Besides the DD-plots, we present the statistical AADDS scores in Tab.~\ref{tab:AADDS-beta-TCVAE}, from which a deteriorating trend can be easily found for latent robustness with the increasing disentangle strength $\beta$. 
	
	
    \small{
    \begin{table*}[tbp]
    \centering
    \caption{AADDS scores for $\beta$-TCVAE models on black-targeted attack experiments. Limit-lines are set as $0.06, 0.75, 3e^{-4}, 34$ for LPIPS, SSIM, FID, and PSNR, respectively.}
    \label{tab:AADDS-beta-TCVAE}
    \begin{tabular}{@{}lllllllll@{}}
    \toprule
    \multicolumn{1}{c}{$\beta$} &   LPIPS       & SSIM          & FID         & PSNR\\ \cmidrule(r){1-5}
    1                           & 2.0859E-02    & 8.4344E-02    & 1.2830E-04  & 9.7208E-01               \\
    5                           & 2.0824E-02    & 8.4787E-02    & 1.1193E-04  & 1.0091E-00               \\
    10                          & 2.0334E-02    & 8.3342E-02    & 1.0375E-04  & 9.4440E-01              \\
    30                          & 1.9575E-02    & 8.1270E-02    & 1.1711E-04  & 8.8342E-01               \\
    50                          & 1.7856E-02    & 7.4351E-02    & 1.2290E-04  & 7.8939E-01    \\ \bottomrule
    \end{tabular}
    \end{table*}
    }

\section{Conclusion and Discussion}\label{conclusion}
	This empirical study investigates robustness issues about the latent space for generative autoencoders. 
	We verify the vulnerability of the variational autoencoders by attacking them from the latent space. 
	Experiments on three types of VAE models trained on MNIST, FasionMNIST, and CelebA datasets show that one can mislead the decoders to reconstruct images quite different from the original or even completely invalid. 
	We also develop the adversarial latent training framework and achieve more robust VAEs. 
	Furthermore, experiments are conducted for latent robust comparison between variational autoencoders and deterministic autoencoders. 
	The results give us a new insight into that the DAEs are more robust in the latent space than VAEs. 
	Finally, we discuss the relationship between disentanglement and the latent robustness of the $\beta$-TC-VAE models. 
	The finding is that the promotion of disentanglement may lead decline in latent robustness.
	To wrap up, we explore several points related to the latent robustness of VAEs, giving certain explanations and insights.
	
	We see two important directions for further research.
	First is the theoretical analysis of the relationship between latent robustness and the generation diversity for VAEs.
    Latent robust generative autoencoders tend to generate more homogenized samples. 
    There may be a trade-off between latent robustness and generation diversity.
	Second, an extension of investigation for adversarial latent robustness in other research areas of artificial intelligence such as natural language processing and so on.
	Experiments in this paper are all conducted on autoencoders for datasets from the field of computer vision. 
	Well, the framework and methodology are easy to extend for networks in applications of other domains. 
\bibliographystyle{elsarticle-num}
\bibliography{LatentRobustness.bib}	




\end{document}